\documentclass[sigconf]{acmart}
\usepackage{xcolor,soul}
\usepackage{hyperref}
\usepackage{colortbl}
\usepackage{graphicx} 
\usepackage{enumitem}
\usepackage{balance}
\usepackage{makecell}
\definecolor{mypink}{rgb}{.99,.90,.92}
\definecolor{magicmint}{rgb}{0.67, 0.94, 0.82}
\usepackage{tcolorbox}
\usepackage{caption}
\usepackage{multicol}
\usepackage{float}
\setcopyright{none}
\renewcommand\footnotetextcopyrightpermission[1]{}
\AtBeginDocument{%
  }

\copyrightyear{2025}
\acmYear{2025}
\setcopyright{cc}
\setcctype{by}
\acmConference[MM '25]{Proceedings of the 33rd ACM International Conference on Multimedia}{October 27--31, 2025}{Dublin, Ireland}
\acmBooktitle{Proceedings of the 33rd ACM International Conference on Multimedia (MM '25), October 27--31, 2025, Dublin, Ireland}\acmDOI{10.1145/3746027.3762070}
\acmISBN{979-8-4007-2035-2/2025/10}

\begin{document}

\title{Structured Prompting and LLM Ensembling for Multimodal Conversational Aspect-based Sentiment Analysis}

\author{Zhiqiang Gao}
\affiliation{
  \institution{Hunan University}
  \city{Changsha}
  \state{Hunan}
  \country{China}
}
\email{gaozhiqiang@hnu.edu.cn}

\author{Shihao Gao}
\affiliation{
  \institution{Hunan University}
  \city{Changsha}
  \state{Hunan}
  \country{China}
}
\email{shihaogao@hnu.edu.cn}

\author{Zixing Zhang}
\affiliation{
  \institution{Hunan University}
  \city{Changsha}
  \state{Hunan}
  \country{China}
}
\email{zixingzhang@hnu.edu.cn}

\author{Yihao Guo}
\affiliation{
  \institution{Hunan University}
  \city{Changsha}
  \state{Hunan}
  \country{China}
}
\email{guoyihao@hnu.edu.cn}

\author{Hongyu Chen}
\affiliation{
  \institution{Hunan University}
  \city{Changsha}
  \state{Hunan}
  \country{China}
}
\email{redtea@hnu.edu.cn}

\author{Jing Han}
\authornote{Jing Han is the corresponding author.}
\affiliation{
  \institution{University of Cambridge}
  \city{Cambridge}
  \country{UK}
}
\email{jh2298@cam.ac.uk}

\renewcommand{\shortauthors}{Gao et al.}

\begin{abstract}
Understanding sentiment in multimodal conversations is a complex yet crucial challenge toward building emotionally intelligent AI systems. The Multimodal Conversational Aspect-based Sentiment Analysis (MCABSA) Challenge invited participants to tackle two demanding subtasks: (1) extracting a comprehensive sentiment sextuple—including holder, target, aspect, opinion, sentiment, and rationale—from multi-speaker dialogues, and (2) detecting sentiment flipping, which detects dynamic sentiment shifts and their underlying triggers.
For Subtask-I, in the present paper, we designed a structured prompting pipeline that guided large language models (LLMs) to sequentially extract sentiment components with refined contextual understanding. For Subtask-II, we further leveraged the complementary strengths of three LLMs through ensembling to robustly identify sentiment transitions and their triggers. Our system achieved a 47.38\% average score on Subtask-I and a 74.12\% exact match F1 on Subtask-II, showing the effectiveness of step-wise refinement and ensemble strategies in rich, multimodal sentiment analysis tasks.
\end{abstract}






\maketitle

\section{Introduction}

Emotional understanding represents a crucial capability for artificial intelligence. This has positioned sentiment analysis and opinion mining as critical research frontiers. Over the past few years, driven by the continuous development of AI technologies, sentiment analysis has achieved significant progress~\cite{medhat2014sentiment, schouten2015survey,wang2025multimodal,chow2024unified,wu2025combating,fei2024video}. The field has evolved from traditional coarse-grained methods, such as document-level and sentence-level analysis, to fine-grained approaches like Aspect-Based Sentiment Analysis (ABSA)~\cite{8976252, ZhangLDBL23, zha2025dual, chen2024gradient,zhang2024exploring}, which incorporates nuanced emotional elements and enables the extraction of complex sentiment tuples, including targets, aspects, opinions, and sentiments~\cite{emotion-cause, fan2021multi,alaei2023target,wan2020target,ma2018targeted}. 
As technology progresses, a growing emphasis has been observed on integrating multimodal signals (i.e., facial expressions, vocal intonations, and gestures) into communication systems to capture subtle emotional cues that go beyond the limitations of text alone~\cite{yang2022face,ben2021video,10089511,zheng2025multi,yan2024modeling,poddar2017author}.
Consequently, sentiment analysis has progressively embraced multimodal approaches that integrate text, speech, images, and video~\cite{zhang2024open, XuDWX025, 10888592, zhang2024re, zheng2023facial, gu2021targeted, li2024multi}.


In line with this progression, a significant step was taken by Luo et al.~\cite{luo2024panosent}. Their work pioneered the formalization of sentiment analysis in multimodal, multi-party conversations by devising a comprehensive six-element framework. Their work also introduced the notion of tracking dynamic sentiment shifts and their underlying causes. To support research in this direction, they released \textit{PanoSent}, a large-scale dataset of 10{,}000 annotated dialogues across three languages, covering over 100 common domains and scenarios. With this, they launched the \textit{Multimodal Conversational Aspect-based Sentiment Analysis (MCABSA)} Challenge at ACM MM 2025, which features two subtasks: \textit{Panoptic Sentiment Sextuple Extraction} and \textit{Sentiment Flipping Analysis}.


The MCABSA challenge presents a primary difficulty. Due to the complexity of directly extracting the six sentiment elements as a single, intricate task, which is difficult for a monolithic model to solve effectively. Inspired by the Chain-of-Thought (CoT)~\cite{wei2022chain} paradigm, prior work has decomposed this complex task into three simpler subtasks~\cite{luo2024panosent}. However, this CoT approach introduces another issue: susceptibility to error propagation. An error in identifying an early-stage element, will inevitably lead to inaccuracies in the subsequent dependent elements.

To address this, we propose a novel framework that leverages \textit{structured prompting}, namely using schema-aligned prompts to guide LLMs in generating accurate and coherent outputs. In particular, our method is characterised by the following three key contributions. First, we introduce a \textbf{Multi-Sampling Generation and Refinement (MSGR)} method for accurate Target-Aspect extraction, which generates diverse candidates using a fine-tuned model and selects the final pair via a powerful LLM to reduce early-stage errors. Second, we introduce a \textbf{Hybrid LLM Optimization Strategy (HLOS)}, where an LLM refines the Opinion, Sentiment, and Rationale components produced by the base model, enhancing overall sextuple prediction. Third, for Subtask-II, we develop an \textbf{ensemble strategy} that combines outputs from three LLMs to improve the robustness and generalization in sentiment flipping analysis.

Our method achieved an average score (Sextuple
Micro F1 Score and Identification F1 Score) of 47.38\% for Subtask-I and an Exact Match F1 Score of 74.12\% for Subtask-II in the final evaluation, securing a third-place ranking among all participating teams.
\section{Challenge Details}
\label{sec:tasks}





\begin{figure}[t]
  \centering
  \includegraphics[width=\linewidth]{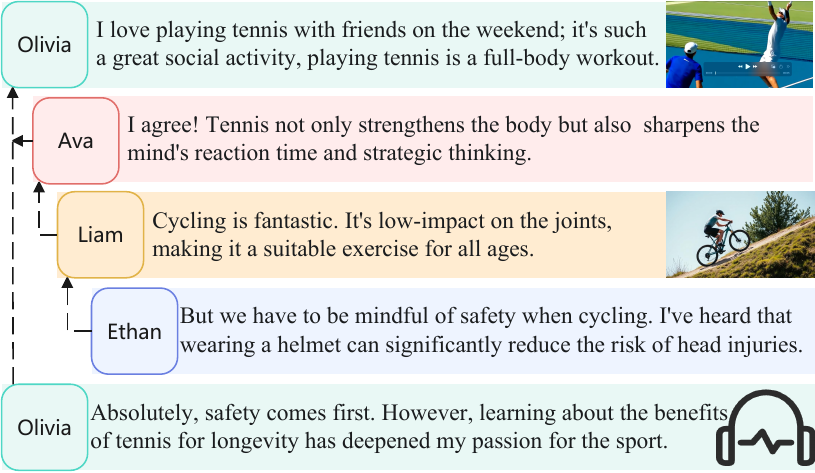}
  \caption{\textbf{Illustration of the input data format provided in the MCABSA Challenge.} The data comprises multi-party, multi-turn dialogues that integrate textual content with multimodal elements, including audio, images, and videos. Speaker information is also included alongside the dialogue content.}
  \label{example}
\end{figure}

The MCABSA Challenge, part of ACM Multimedia 2025, builds on the PanoSent dataset~\cite{luo2024panosent}. An example data point from this dataset is illustrated in Figure~\ref{example}. The challenge features two subtasks, which we describe in the following.

\noindent\textbf{Subtask-I: Sentiment Sextuple Extraction.}
This subtask requires the extraction of a comprehensive sentiment sextuple: \textit{(holder, target, aspect, opinion, sentiment, rationale)} from multi-turn, multi-party, and multimodal dialogues. The elements can be explicit text spans or implicit concepts that must be inferred from the conversational context and non-textual modalities (image, audio, video).

\textit{Dataset.} The train set provided to participants comprises 5,600 English-language dialogues, encompassing 27,100 utterances and involving 25,300 speakers. Among these, 3,400 dialogues consist solely of textual content, while the remaining 2,200 incorporate one or more multimodal elements (e.g., image, audio, or video). The test set includes 1,500 dialogues, on which participants are required to predict sentiment sextuples for evaluation and final ranking.


\textit{Evaluation and Ranking.} 
Performance is evaluated using two metrics. The first, \textit{Sextuple Micro F1 Score}, measures the accuracy of predicting the complete sextuple and is defined as:
\begin{equation}
  \text{Micro F1} = 2 \times \frac{OP \times OR}{OP + OR},
\end{equation}
where
\begin{equation}
  OP = \frac{\text{\# correctly predicted sextuples}}{\text{\# predicted sextuples}},
\end{equation}
and
\begin{equation}
  OR = \frac{\text{\# correctly predicted sextuples}}{\text{\# gold sextuples}}.
\end{equation}
The second metric, \textit{Identification F1 Score}, which computes the same score excluding sentiment polarity.
Final ranking is based on the average of the two, and higher scores indicates better performance.

\begin{figure*}[htp]
    \centering
    \includegraphics[width=\textwidth]{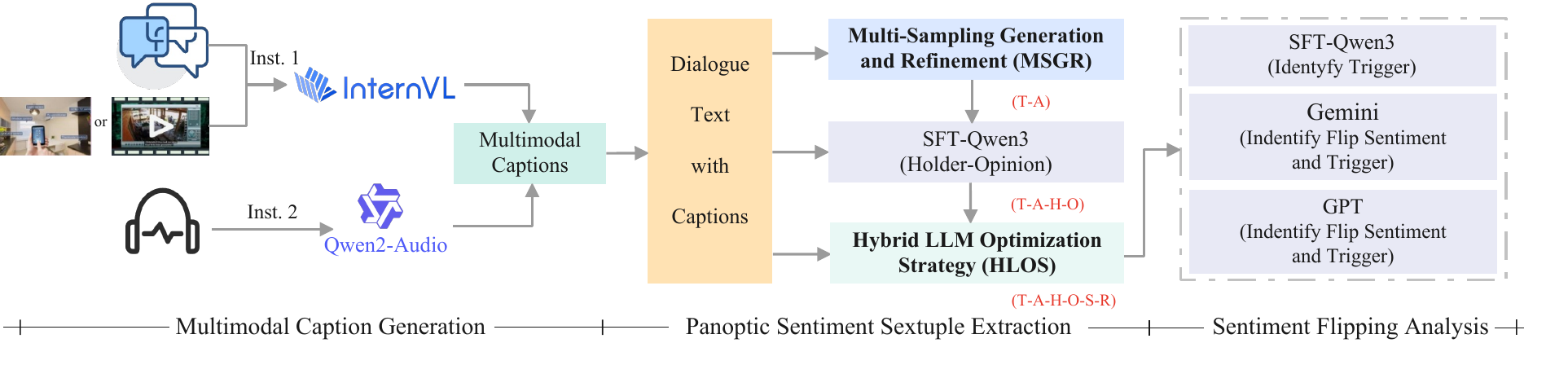}
    \caption{\textbf{Overview of the proposed framework for the MCABSA Challenge.} The framework consists of three main components: (1) \textit{multimodal caption generation}, where audio, image, and video descriptions are extracted and integrated into the dialogue input; (2) \textit{Panoptic Sentiment Sextuple Extraction}, which performs step-wise extraction of sentiment sextuples; and (3) \textit{Sentiment Flipping Analysis}, which analyzes sentiment flipping through an ensemble approach with three models.
    }
    \label{Overview}
\end{figure*}

\noindent\textbf{Subtask-II: Sentiment Flipping Analysis.}
%
This subtask is to detect dynamic changes in a speaker's sentiment towards the same target-aspect pair within a dialogue. The goal is to identify instances of sentiment flips and classify their cause, outputting a tuple of \textit{(holder, target, aspect, initial sentiment, flipped sentiment, trigger)}. The trigger must be classified into one of four predefined categories: \textit{introduction of new information}, \textit{logical argumentation}, \textit{participant feedback and interaction}, or \textit{personal experience and self-reflection}. 

\textit{Dataset.} The dataset provided to participants for model training in this subtask consists of 1,754 dialogues, representing a subset of the data from Subtask-I. All selected dialogues exhibit distinct sentiment flips. Notably, the dataset does not include a predefined split for training and development sets. The test set comprises 520 dialogues, also drawn as a subset from the Subtask-I test data.


\textit{Evaluation and Ranking.}  
Submissions are evaluated using the \textit{Exact Match F1 Score}, which measures how well the model simultaneously identifies the correct flipped sentiment and its trigger. Final rankings are determined directly by this score, with higher values indicating better performance. 
All metrics are computed using Python scripts provided by the challenge organizers.

\section{Method}
\label{sec:method}



This section outlines our proposed framework for the MCABSA Challenge, highlighting how each component is designed to tackle the task’s unique challenges.

\subsection{Overvall Architecture}

As illustrated in Figure~\ref{Overview}, the architecture comprises three primary components: multimodal caption generation, sentiment sextuple extraction for \textit{Subtask-I}, and ensemble-based sentiment flipping analysis for \textit{Subtask-II}. The input to our system is a multi-turn dialogue with a reply structure, where each utterance may contain textual content alongside non-verbal modalities such as image, audio, or video. Speaker information is also provided to help contextualize interactions.




To enable a unified understanding of diverse modalities of the input dialogue, we begin with \textit{multimodal caption generation}, with specialized large multimodal models being used to convert non-textual inputs into descriptive textual representations.
These generated captions are then embedded into the corresponding dialogue turns, enriching textual context and enabling unified multimodal analysis.
The full methodology of this stage is described in Section~\ref{sec:stage1}.


We then move to the first core task: \textit{panoptic sentiment sextuple extraction} through a structured three-stage pipeline. More precisely, we deployed three substeps to extract the complete sentiment sextuple: MSGR for \textit{Target-Aspect} (T-A), a fine-tuned LLM for \textit{Holder-Opinion} (H-O), and HLOS for \textit{Sentiment-Rationale} (S-R), with more details provided in Section~\ref{sec:stage2}.


Using these extracted sentiment sextuples as a foundation, Subtask-II focuses on \textit{identifying sentiment shifts}.
For this aim, we adopt an ensemble strategy that combines three complementary models. 
The outputs from all models are then aggregated to produce the final predictions. Further details are provided in Section~\ref{sec:stage3}.


\begin{figure}[tp]
    \centering
    \includegraphics[width=0.5\textwidth]{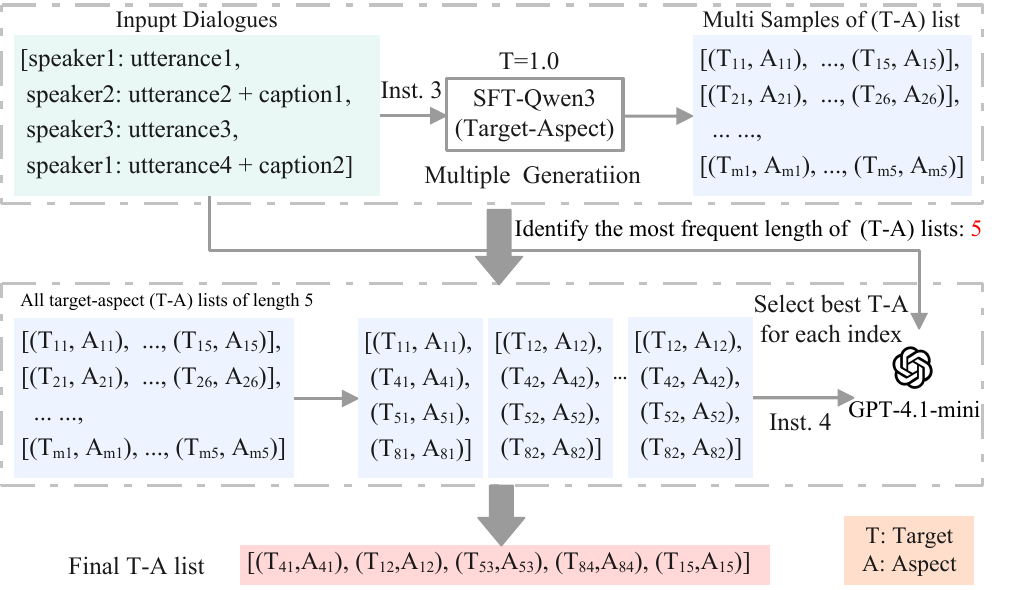}
    \caption{\textbf{Multi-Sampling Generation and Refinement (MSGR) for target-aspect extraction.}}
    \label{MSGR}
\end{figure}

\subsection{Multimodal Caption Generation}
\label{sec:stage1}


The input comprises multi-party, multi-turn dialogues enriched with occasional multimodal elements such as audio, image, and video~\cite{luo2024panosent}. As these modalities often convey key sentiment cues for Subtask-I, we convert them into textual descriptions to enable a unified, text-based processing pipeline.

As depicted in Figure~\ref{Overview}, for audio inputs, we leverage Qwen2-Audio to perform Automatic Speech Recognition, directly transcribing the audio content into textual form. For images and videos, we leverage the strong captioning capabilities of modern multimodal LLMs. Specifically, InternVL-3-14B is prompted with both the dialogue context and visual input to generate descriptions that align with the surrounding conversation. This zero-shot or few-shot prompting strategy~\cite{few-shot} enables efficient and context-aware caption generation with minimal computational cost.

\subsection{Multi-Sampling Generation and Refinement Method for Target and Aspect Extraction}
\label{sec:stage2}
To effectively extract targets and aspects from dialogues, we propose the Multi-Sampling Generation and Refinement Method. This approach addresses the inherent variability in LLM outputs by leveraging stochastic sampling and consensus-based refinement, ensuring more robust and accurate identification of T-A pairs. As illustrated in Figure~\ref{MSGR}, MSGR comprises two main steps: (1) determining the optimal length of the T-A list, and (2) refining the specific T-A elements at each position.

In the first step, we focus on determining the length of the T-A list for a given dialogue. To introduce diversity in the generations, we employ a fine-tuned Qwen3-8B with a relatively high temperature coefficient (i.e., T=1) during inference. This setup encourages variability in the model's outputs. For each dialogue, we perform multiple inference runs (i.e., repeated generations) on the same input prompt, yielding a set of T-A lists with potentially differing lengths. We define a threshold \( h \) to establish consensus: for each generated list, we record its length and maintain a frequency count across all generations. The process continues until the frequency of a particular length reaches \( h \) first, at which point that length \( n \)
is selected as the final T-A list length for the dialogue. This majority-voting mechanism mitigates the impact of outlier generations and stabilizes the extraction process.

The second step refines the actual T-A pairs based on the determined length \( n \) and the \( h \) candidate lists that share this length. For each position \( i \) (where \( 0 \leq i < n \)), we aggregate the T-A pairs at index \( i \) from the \( h \) lists, forming a candidate set for that position. This set is then fed, along with the original dialogue context, to GPT-4.1-mini~\cite{GPT4.1} via a crafted prompt. The prompt instructs the model to select the most appropriate T-A pair from the candidates, considering contextual relevance, consistency with the dialogue, and alignment with multimodal cues (if applicable). This selection is performed independently for each index, resulting in a refined T-A list of length \( n \). 
MSGR generates diverse extraction candidates and employs a secondary LLM for adjudication, enhancing robustness against hallucinations. 

\subsection{Hybrid LLM Optimization Strategy for Enhanced Sentiment Sextuple Extraction}
\label{sec:stage3}
\begin{figure}[t]
    \centering
    \includegraphics[width=0.37\textwidth]{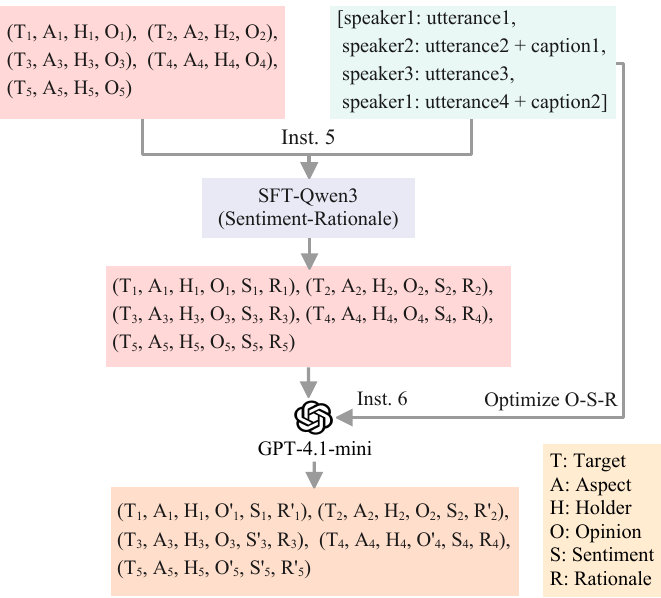}
    \caption{\textbf{Hybrid LLM Optimization Strategy (HLOS) for enhanced sentiment sextuple extraction.}} 
    \label{HLOS}
\end{figure}

To refine the opinion, sentiment, and rationale components within the sextuples, we introduce HLOS. This method combines the strengths of fine-tuned models for initial extraction with advanced prompting of more capable LLMs for verification and correction, thereby improving the accuracy and contextual alignment of the extracted elements. As shown in Figure~\ref{HLOS}, HLOS operates in two sequential steps: (1) initial completion of the sextuple using a fine-tuned model, and (2) subsequent refinement via a 2nd LLM.

In the first step, we input the dialogue context along with the pre-extracted elements—target, aspect, holder, and opinion—to a fine-tuned Qwen3-8B model. This model has been specifically fine-tuned with instructions tailored for extracting sentiment and rationale. The output of this step is a preliminary complete sextuple, providing a structured foundation for further optimization.

The second step employs GPT-4.1-mini, 
to perform a targeted review and revision. Through carefully designed prompts, we instruct the model to evaluate the initial sextuple against the full dialogue context, including any multimodal captions. The focus is on assessing and correcting the opinion, sentiment, and rationale for consistency, logical coherence, and alignment with the conversational dynamics. For instance, the prompt guides the model to identify potential mismatches (e.g., inferred rationales not supported by context) and propose revisions, outputting an optimized sextuple. This hybrid approach leverages the efficiency of fine-tuning for initial predictions while harnessing the superior reasoning capabilities of GPT-4.1-mini for high-fidelity refinements.

\subsection{Integrated Model Approach for Sentiment Flipping Analysis}

For Subtask-II, we propose an integrated ensemble approach to detect sentiment shifts and identify their underlying triggers within multi-turn dialogues. The subtask can be further divided into two phases: (1) detecting whether a sentiment flip occurs for a given holder–target–aspect triple, and (2) classifying the trigger responsible for the flip. To improve robustness and accuracy, we combine three complementary models and leverage Subtask-I outputs to reduce redundancy.


Specifically, the first model builds on the sentiment sextuples extracted in Subtask-I. We apply rule-based logic to track the sentiment trajectory expressed by the same holder toward the same target–aspect pair across turns. A flip is flagged when a sentiment change is detected. Once identified, a \textit{fine-tuned Qwen3-8B} model classifies the trigger using context-aware prompts that incorporate both dialogue and multimodal captions.
The second model employs \textit{Gemini-2.5-pro} in a zero-shot prompting fashion for end-to-end flip detection and trigger classification. This model processes the full dialogue context—including multimodal inputs—without relying on intermediate Subtask-I outputs. Similarly, the third model uses \textit{GPT-4.1} via API with tailored prompts to perform the same task.


In our observations of the individual model performances, Gemini-2.5-pro achieves the best results, followed by GPT-4.1 and the fine-tuned Qwen3-8B. However, since the top two models occasionally produce empty outputs, we implement a hierarchical fusion strategy to address this: Gemini-2.5-pro serves as the base model; when its outputs are empty, we backfill with GPT-4.1, and then with Qwen3-8B as a fallback. 
This hierarchical fusion strategy prioritizes high-performing models while filling gaps from lower-tier ones to ensure completeness. 



\section{Experiments and Results}
\label{sec:exper}

\noindent\textbf{{Experimental Setup.}}
All experiments were conducted on a server equipped with eight NVIDIA RTX 3090 GPUs. For Subtask-I, we used Qwen2-Audio to generate audio captions and InternVL-3-14B for image and video captions. The Qwen3-8B model was fine-tuned for sextuple extraction. MSGR and HLOS methods were applied, with GPT-4.1-mini accessed via the OpenAI API for refinement. For Subtask-II, Qwen3-8B was fine-tuned, while GPT-4.1 and Gemini-2.5-pro were utilized through their respective APIs.


For both subtasks, Qwen3-8B fine-tuning was performed using the SWIFT~\cite{zhao2025swift} framework with Low-Rank Adaptation (LoRA). In particular, the model was fine-tuned using bfloat16 precision over 2 epochs with a batch size of 2. Flash Attention was enabled to optimize memory and computation efficiency. The learning rate was set to 1e-5. LoRA parameters were configured with a rank of 8 and an alpha of 32. The maximum input sequence length was limited to 2\,048 tokens. 
We reserved 10\% of the training data as a validation split (randomly selected) to enable internal evaluation before official test submission.

\begin{table}[t] 
\centering
\setlength{\tabcolsep}{6pt}
\caption{\textbf{Performance comparison of various methods on Subtask-I using average score of Sextuple Micro F1 Score and Identification F1 Score on the validation and test sets.
}}
\label{Subtask_1_results}
\begin{tabular}{c|c|ccc}
\toprule
\textbf{Method}                    & \textbf{Val.}         & \textbf{Test}          \\
\midrule
Sentica (+CoT)           & 28.82              & /             \\
Sentica (+CoS)           & 31.70              & /             \\
Sentica (+CoS+PpV)       & 33.95              & /             \\
\midrule
SFT-Qwen3-8B             & 35.61              &  44.52        \\
MSGR+SFT-Qwen3-8B        & 37.24              &  44.97        \\
SFT-Qwen3-8B+HLOS        & 43.59              &  47.30        \\
\midrule
\textit{\textbf{MSGR+SFT-Qwen3-8B+HLOS}}   & \textbf{44.37}              &  \textbf{47.38}        \\
\bottomrule
\end{tabular}
\end{table}

\noindent\textbf{{Results of Subtask-I.}}
\label{sec:result-task1}
We evaluate our framework on both the validation split and the official test set, benchmarking against the challenge baseline (Sentica variants) and conducting ablations on key components: fine-tuned Qwen3-8B for sextuple extraction, MSGR for target-aspect (T-A) identification, and HLOS for opinion-sentiment-rationale (O-S-R) refinement.


Table~\ref{Subtask_1_results} presents the results. On the validation split, the baseline achieves 33.95\%. Applying a three-step pipeline with fine-tuned Qwen3-8B (extracting T-A, holder-opinion (H-O), and O-S-R) raises performance to 35.61\%, showing the benefit of modular fine-tuning. Replacing the T-A step with MSGR further improves results to 37.24\%, highlighting MSGR’s strength in capturing T-A elements through multi-sampling and refinement. Introducing HLOS on top of the Qwen3-8B outputs leads to a significant jump to 43.59\%, demonstrating the effectiveness of LLM-based refinement. Our full pipeline—integrating MSGR, Qwen3-8B, and HLOS—achieves the best result at 44.37\%, with each component contributing incremental gains.
On the test set, the full pipeline achieves 47.38\%, with consistent gains across ablations confirming its robustness. Notably, HLOS provides the largest single uplift, while MSGR contributes additional improvements in T-A extraction. These findings highlight the complementary strengths of each module and the framework’s ability to generalize well to unseen, multimodal data.



\begin{table}[t] 
\centering
\setlength{\tabcolsep}{6pt}
\caption{\textbf{Performance comparison of various methods on Subtask-II using Exact Match F1 scores on the validation and test sets.} Methods include baselines, and the introduced fine-tuned Qwen3-8B, GPT-4.1, Gemini-2.5-pro, and the ensemble fusions of multiple results.}
\label{Subtak_2_results}
\begin{tabular}{c|c|ccc}
\toprule
\textbf{Method}                    & \textbf{Val.}         & \textbf{Test}          \\
\midrule
Sentica (+CoT)         & 55.28              & /             \\
Sentica (+CoS)         & 59.40              & /             \\
Sentica (+CoS+PpV)     & 62.52              & /             \\
\midrule
SFT-Qwen3-8B           & 65.89              & 60.28         \\
GPT-4.1                & 84.11              & 71.73         \\
Gemini-2.5-pro         & 84.28              & 73.18         \\
\midrule
\textit{\textbf{Gemini-2.5-pro+GPT-4.1}} & \textbf{86.25}              & \textbf{74.12}         \\
\bottomrule
\end{tabular}
\end{table}


\noindent\textbf{{Results of Subtask-II.}}
\label{sec:result-task2}
Table~\ref{Subtak_2_results} presents the results for Sentiment Flipping Analysis on both validation split and the official test set. The baseline model achieves a modest 62.52\%. Our rule-based approach combined with fine-tuned Qwen3-8B improves this to 65.89\%.
More substantial gains come from direct end-to-end prompting with stronger proprietary models: GPT-4.1 achieves 84.11\%, and Gemini-2.5-pro slightly surpasses it at 84.28\%. Our ensemble method, which integrates all three models using a hierarchical fallback strategy, achieves the highest performance at 86.25\%. This implies the robustness and complementarity of the fusion design.
On the test set, all methods experience a performance drop, likely due to distributional shifts or more complex multimodal contexts. Despite this, the ensemble remains the top performer with 74.12\%, again outperforming all individual models and demonstrating its effectiveness.

\begin{figure*}[t]
  \centering
  \includegraphics[width=.9\linewidth]{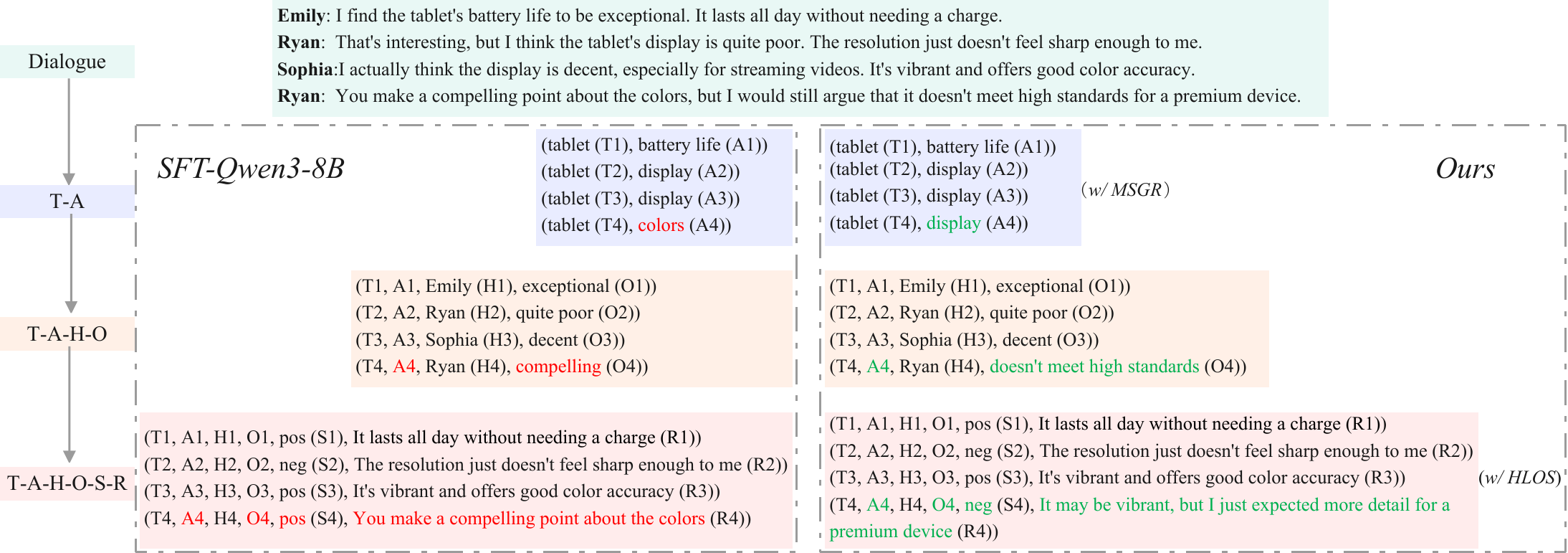}
  \caption{\textbf{A representative example illustrating improvements from MSGR and HLOS over the baseline on Subtask-I. } Results in red indicate errors made by the baseline (left), while those in green show the corrected outputs produced by our method (right).}
  \label{fig:case_study}
\end{figure*}

Additionally, to further illustrate the effectiveness of our proposed approach, Figure~\ref{fig:case_study} presents a representative example from the challenge dataset. In this case, MSGR successfully identifies the correct target-aspect (T-A) pairs, while HLOS refines the sentiment and rationale, resulting in an accurate sentiment sextuple for Subtask-I. This correction also prevents an incorrect sentiment flip detection in Subtask-II.

\section{Discussion}
\label{sec:diss}


Throughout our framework, a pivotal insight emerges from the application of the HLOS in Subtask-I , which consistently yields substantial performance gains on both the validation and test sets. The core rationale underlying this improvement lies in strategically assigning tasks to models based on their inherent strengths, thereby optimizing the overall extraction pipeline.

We experimented with directly prompting GPT-4.1-mini to generate sentiment sextuples from dialogue content alone. Despite its formidable reasoning and comprehension capabilities, this approach underperformed. The primary challenge stems from the diverse and often concise nature of targets and aspects—typically comprising single words or short phrases with high variability across dialogues. Few-shot prompting proved insufficient to enhance the generalization of GPT-4.1-mini for these elements, resulting in suboptimal T-A extraction. Such initial inaccuracies propagated errors, adversely impacting the downstream identification of holders, opinions, sentiments, and rationales. In contrast, the fine-tuned Qwen3-8B model excels in T-A extraction due to its exposure to a vast array of examples during supervised fine-tuning, enabling robust pattern recognition and handling of distributional diversity. However, this model frequently falls short in capturing opinions and rationales, which are typically structured as short sentences or phrases, leading to incomplete or fragmented outputs.

HLOS addresses these limitations by leveraging GPT-4.1-mini for post-extraction refinement. By providing the model with a preliminary complete sextuple as a reference—derived from the outputs of fine-tuned Qwen3-8B, GPT-4.1-mini can more effectively harness its analytical prowess. This referential framework facilitates targeted corrections, particularly for opinion, sentiment, and rationale, ensuring contextual coherence and completeness. Consequently, HLOS achieves marked performance elevations, as evidenced by the empirical results.


\textbf{Limitations.} Although our proposed MSGR aims to enhance T-A extraction as much as possible, its contribution to overall performance improvements remains limited. This can be attributed to the inherent challenges in capturing the full diversity and subtlety of T-A pairs in multimodal dialogues, where variability in contextual cues and multimodal integrations may not be fully mitigated by sampling alone. Furthermore, our hierarchical approach to sextuple extraction introduces strong inter-component dependencies, which unavoidably propagate errors through the pipeline. For instance, inaccuracies in early stages can cascade to downstream elements like opinion, sentiment, and rationale, amplifying overall inconsistencies despite modular refinements. The inclusion of fine-tuned LLM, such as Qwen3-8B, also poses challenges to the transferability of system. Fine-tuning on specific datasets may limit generalization to new domains or languages, requiring additional adaptation efforts for broader applicability. Additionally, the repeated utilization of proprietary models like GPT-4.1-mini throughout the framework—particularly in refinement and ensemble stages—increases computational costs, potentially hindering scalability in resource-constrained environments. Future iterations could explore open-source alternatives or optimization techniques to address these drawbacks.


\textbf{Lessons and Future Work.} To address error propagation inherent in the hierarchical pipeline, one key avenue is the development of end-to-end models tailored for dialogue sentiment analysis. Such models could jointly optimize all sextuple elements, potentially incorporating attention mechanisms or transformer-based architectures that process multimodal inputs holistically, thereby minimizing cascading inaccuracies and improving overall coherence.

Additionally, constrained by computational resources and time in our experiments, we did not explore a broader range of models. Future work could involve substituting the open-source components, such as Qwen3-8B, with larger-scale open-source LLMs to leverage enhanced capacity for handling complex dialogues and multimodal integrations, while maintaining accessibility and reproducibility.

For Subtask-II, incorporating relational information among speakers could augment LLM-based trigger classification. This might involve graph-based representations of dialogue structures to assist models in discerning nuanced flip causes, thereby boosting accuracy in detecting and categorizing sentiment shifts. Finally, the limited scale and moderate quality of the Subtask-II training data present opportunities for data-centric improvements. Expanding the dataset through augmentation techniques, crowdsourcing high-quality annotations, or synthesizing realistic multimodal dialogues might elevate benchmark standards in the MCABSA domain, fostering more generalizable models and advancing the field toward real-world applications.
\section{Conclusions}
\label{sec:conc}



This paper presents a framework we developed for the MCABSA Challenge, targeting Panoptic Sentiment Sextuple Extraction and Sentiment Flipping Analysis in multimodal dialogues. 
Our proposed system combines multimodal caption generation, a Multi-Sampling Generation and Refinement Method for Target-Aspect extraction, a Hybrid LLM Optimization Strategy for refining opinion, sentiment, and rationale, and an ensemble model for sentiment flipping detection. 
In the final evaluation, our approach achieved scores of 47.38\% and 74.12\% for the two subtasks, respectively. 
The results demonstrate the feasibility of the proposed approach and offer a foundation for further work on end-to-end multimodal sentiment understanding. 
Future work includes leveraging larger open-source LLMs for improved multimodal reasoning~\cite{Multimodalhumanai, wu2024next}, augmenting Subtask-II training data to enhance generalization~\cite{patel2024datadreamer,LongWXZDCW24}, and incorporating relational information via graph-based dialogue representations to better capture speaker dynamics and sentiment shifts~\cite{yin2024textgt,su2024cgt}.


\begin{acks}
This work was mainly supported by the Guangdong Basic and Applied Basic Research Foundation under Grant No.~2024A1515010112 and the Changsha Science and Technology Bureau Foundation under Grant No.~kq2402082. 
We would also like to thank the organisers of the MCABSA challenge for providing the dataset, evaluation platform, and helpful communication throughout the competition.
\end{acks}

\clearpage
\balance

\bibliographystyle{ACM-Reference-Format}
\bibliography{ref}

\end{document}